\documentclass[runningheads]{llncs}
\usepackage{graphicx}
\usepackage{amsmath,amssymb} % define this before the line numbering.
\usepackage{color}
\usepackage[width=122mm,left=12mm,paperwidth=146mm,height=193mm,top=12mm,paperheight=217mm]{geometry}
\usepackage{caption}
\usepackage{algorithm}
\usepackage[noend]{algpseudocode}
\floatname{algorithm}{ }
\usepackage{eqparbox}

\usepackage{times}
\usepackage{epsfig}
\usepackage{siunitx}
\usepackage{makecell}
\def\ie{{i.e.\ }}

\begin{document}
% \linenumbers
\pagestyle{headings}
\mainmatter
%\def\ECCV18SubNumber{201}  % Insert your submission number here

%\title{LTNN: Conditional Transformation Generative Adversarial Network for Image View and Attribute Modification}
\title{Latent Transformations for Object View Points Synthesis}

%\titlerunning{ECCV-18 submission ID \ECCV18SubNumber}

%\authorrunning{ECCV-18 submission ID \ECCV18SubNumber}

%\author{Anonymous ECCV submission}
%\institute{Paper ID \ECCV18SubNumber}

\author{Sangpil Kim, Nick Winovich, Guang Lin, Karthik Ramani}
\institute{Purdue University,
▸ West Lafayette,
  USA\\
}
%▸ \email{ \{kim2030,nwinovic,guanglin,ramani\}@purdue.edu}

%\orcid{1234-5678-9012-3456}
%\affiliation{%
%  \institution{Purdue University}
%  %\streetaddress{Purdue Mall}
%  \city{West Lafayette}
%  \state{IN}}
%  %\postcode{47906}}
%  %\country{USA}}
%\email{kim2030@purdue.edu}
%
%\author{Nick Winovich}
%\orcid{1234-5678-9012-3456}
%\affiliation{%
%  \department{Department of Mathematics}
%  \institution{Purdue University}
%  \city{West Lafayette}
%  \state{IN}}
%\email{nwinovic@purdue.edu}

\maketitle
\begin{abstract}
  We propose a fully-convolutional conditional generative model, the latent transformation neural network (LTNN),
  capable of view synthesis using a light-weight neural network suited for real-time applications.
  In contrast to existing conditional generative models
  which incorporate conditioning information via concatenation,
  we introduce a dedicated network component, the conditional transformation unit (CTU), designed to learn the latent space transformations corresponding to specified target views.
  In addition, a consistency loss term is defined to guide the network toward learning the desired latent space mappings,
  a task-divided decoder is constructed to refine the quality of generated views,
  and an adaptive discriminator is introduced to improve the adversarial training process.
  The generality of the proposed methodology is demonstrated on a collection of three diverse tasks: multi-view reconstruction on real hand depth images, view synthesis of real and synthetic faces, and the rotation of rigid objects.
  The proposed model is shown to exceed state-of-the-art results in each category while simultaneously achieving
  a reduction in the computational demand required for inference by 30\% on average.
\end{abstract}

\section{Introduction}
%%  ---  ATTEMPT 3  ---
%%
Generative models have been shown to provide %exceptionally
effective frameworks for
representing complex, structured datasets and generating realistic samples from underlying data distributions~\cite{Goodfellow17}.
%%
%This concept has been extended to that of conditional generative models
This concept has also been extended to form conditional models
capable of sampling from conditional distributions %, thus allowing for
%certain properties of the generated data to be controlled.
in order to allow certain properties of the generated data to be controlled or selected~\cite{mirza2014conditional}.
These generative models are designed to sample from broad classes of the
data distribution, %~\cite{van2016conditional},  %(``Man'' or ``Woman'')
however, and are not suitable for
inference tasks which require identity preservation of the input data. % (``John'' or ``Jane'').
%%
%Several
Models have also been proposed which incorporate encoding components to overcome this
by learning to map input data to an associated
{\emph{latent space}} representation within a generative framework~\cite{makhzani2015adversarial}.
The resulting inference models allow for the defining structure/features of inputs to
be preserved while specified target properties are adjusted through conditioning~\cite{yan2016attribute2image}. % ~\cite{sohn2015learning}.
%while maintaining the
%  tasks such as view synthesis.
Conventional conditional models have largely relied on rather simple methods, such as concatenation,
%for the inclusion of condititioning information;
for implementing this conditioning process;
however, ~\cite{miyato2018cgans} have shown that utilizing the
conditioning information in a less trivial, more methodical manner
has the potential to significantly improve the performance of conditional generative models.
%cGANs with projection discriminators have
%recently been shown to achieve significantly better performance by
%introducing the conditioning information in a less trivial, more methodical manner~\cite{miyato2018cgans}.
%%
In this work, we %introduce the latent transformation neural network (LTNN) and
provide a general framework for effectively performing inference with conditional generative models
by strategically controlling the interaction between conditioning information and latent representations
within a generative inference model. %underlying network structure.
%%
%The latent transformation neural network (LTNN) defined in this work
%aims to
%%

%% INTRO NOTES
\iffalse
\begin{itemize}
\item General intro sentence on success of VAE/GAN networks
\item Conditional generative models ability to generate realistic images in controlled manner
\item Motivation for inference based methods (generating from a specific starting image)
\item Example applications (e.g. robotics, avoiding occlusion, etc.)
\item Need for scalable, real-time conditional structure (avoid dense layers/concatenation)
\item Possible mention of cGANs with Projection discriminators to support claim
\item State goal of proposed LTNN (based on above points)
\end{itemize}
\vspace{0.1in}
\fi

%%
%View synthesis of images given specified view points has same functionality as condtional generative neural network (CGNN)
%, which have been designed to leverage the latent space structure to generate data of a specified class, or data possessing certain attributes, based on conditioning information passed to the network~\cite{sohn2015learning, kingma2014semi}.} %kulkarni2015deep, dosovitskiy2015learning}.
%%
%The latent transformation neural network (LTNN) proposed in this work has been designed to address this gap, and is shown to outperform existing methods.

\begin{figure}[t]
  \begin{center}
    \includegraphics[width=0.7\linewidth]{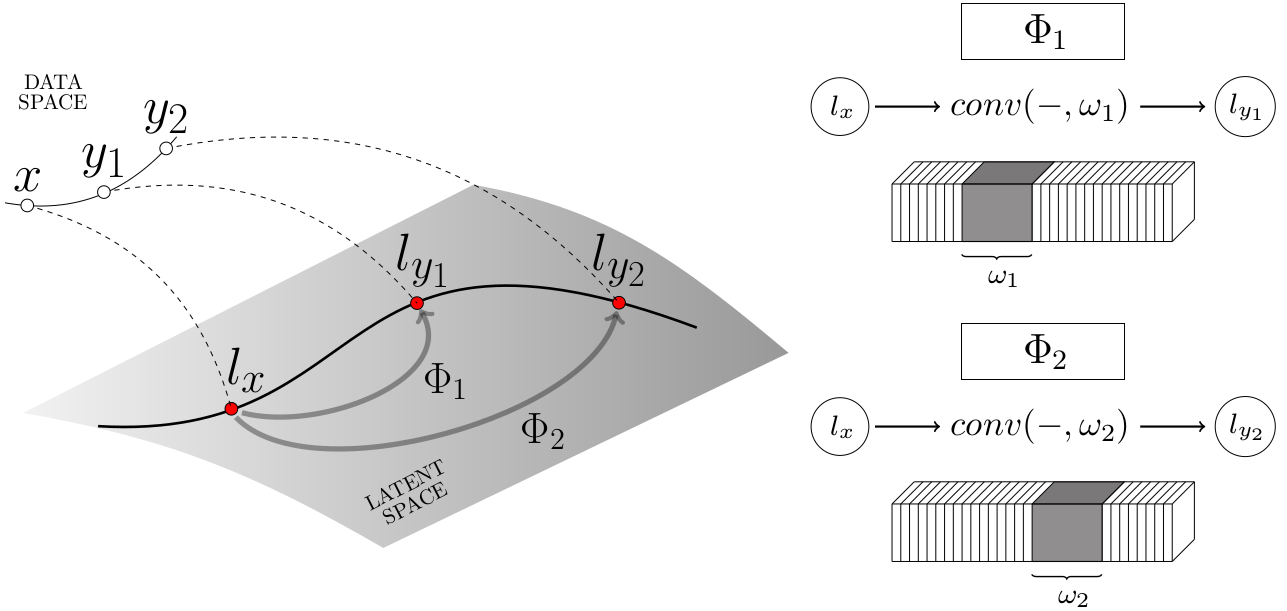}
  \end{center}
  \caption{The conditional transformation unit $\Phi$ constructs a collection of mappings $\{\Phi_k\}$ in the latent space which produce high-level attribute changes to the decoded outputs.
    Conditioning information is used to select the appropriate convolutional weights $\omega_k$ for the specified transformation;
    the encoding $l_x$ of the original input image $x$ is transformed to $\, \widehat{l}_{y_k} = \Phi_k(l_x) = \operatorname{conv}(l_x,\omega_k)$ and provides an approximation to the encoding $\, l_{y_k}$ of the attribute-modified target image $y_k$.}
  \label{fig:manifold}
\end{figure}

%%
In this framework, a conditional transformation unit (CTU), $\Phi$, is introduced to provide a means for navigating the underlying manifold structure of the latent space.  %information encoded in the latent space.
The CTU is realized in the form of a collection of convolutional layers which are designed to approximate the latent space operators defined by mapping encoded inputs to the encoded representations % corresponding to
of specified targets (see Figure \ref{fig:manifold}).
This is enforced by introducing a {\emph{consistency loss}} term %erm in the network's loss function
to guide the CTU mappings during training.
In addition, a conditional discriminator unit (CDU), $\Psi$,  also realized as a collection of convolutional layers, is included in the network's discriminator.
This CDU is designed
to improve the network's ability to identify and eliminate transformation specific artifacts in the network's predictions. %thus improving the overall adversarial training procedure.

The network has also been equipped with RGB balance parameters %$\theta_{RGB}$
consisting of three values $\{\theta_R, \theta_G, \theta_B \}$ designed to give the network the ability to quickly adjust the global color balance of the images it produces to better align with that of the true data distribution.
In this way, the network is easily able to remove unnatural hues and focus on estimating local pixel values by adjusting the three RGB parameters rather than correcting each pixel individually.
%In addition to the RGB balance parameters,
In addition, we introduce a novel estimation strategy for efficiently learning shape and color properties simultaneously; a {\emph{task-divided}} decoder is designed to produce a coarse pixel-value map along with a refinement map in order to split the network's overall task into distinct, dedicated network components. \\

%\newpage
Summary of contributions:
\begin{enumerate}
\item We introduce the conditional transformation unit, with a family of modular filter weights, to learn high-level mappings within a low-dimensional latent space.  In addition, we present a consistency loss term which is used to guide the transformations learned during training.  %; gradient flows are controlled systematically by Swish activation function featuring learnable parameters.
\item We propose a novel framework for color inference which separates the generative process into three distinct network components dedicated to learning (i) coarse pixel value estimates, (ii) pixel refinement scaling factors, and (iii) the global RGB color balance of the dataset.%balance properties. % of the data set.%learning pixel locations, pixel values, and overall/non-local RGB balance into distinct network components dedicated to each task.
\item We introduce the conditional discriminator unit designed to improve adversarial training by identifying and eliminating transformation-specific artifacts present in generated images.      %[Conditional discriminator units]
\end{enumerate}

\iffalse
{\bd{[REPHRASE] We have shown
    the model's ability by improving accuracy of estimation of complex hand poses
    with inferred occluded part of hand from predicted multi-view depth images
    from a single depth image using %both
    real and synthetic data;
    it is shown that multiple, distinct views can be generated simultaneously in real-time.
    In addition,
    the proposed framework significantly outperforms other state-of-the-art conditional generative models at performing simultaneous colorization and attribute-modification on face and chair datasets, producing sharp, realistic predictions.}}
\fi

Each contribution proposed above has been shown to provide a significant improvement to the network's overall performance through a series of ablation studies.
The resulting latent transformation neural network (LTNN)
is placed through a series of comparative studies on a diverse range of experiments
%The resulting network, LTNN,
%is then placed through a series of comparative studies on a diverse range of experiments
where it is seen to outperform existing state-of-the-art models for
(i) simultaneous multi-view reconstruction of real hand depth images in real-time,
(ii) view synthesis and attribute modification of real and synthetic faces,
and (iii) the synthesis of rotated views of rigid objects. % from a single image.
%%
%The light-weight architecture allows the network to synthesize several, distinct views of hand depth images simultaneously in real-time.
%%
%%
%{\rd{
%    The network is also shown to be capable of accurately modeling several discrete modifications simultaneously
%    and can produce seamless continuous attribute modification via piece-wise interpolation.
%}}
%%
%{\rd{
    Moreover, the CTU conditioning framework allows for additional conditioning information, or target views, to be %included into
    added to the training procedure {\emph{ad infinitum}} without any increase to the network's inference speed.
%}}
%%

\section{Related Work}
\cite{dosovitskiy2015learning} has proposed a supervised, conditional generative model trained to generate images of chairs, tables, and cars with specified attributes which are controlled by transformation and view parameters passed to the network.
The range of objects which can be synthesized using the framework is strictly limited to the  pre-defined models used for training; the network can generate different views of these models, but cannot generalize to unseen objects to perform inference tasks.
%% GEOMETRIC PREDICTION
Conditional generative models have been widely used %in computer vision areas such as
for geometric prediction~\cite{park2017transformation,tatarchenko2016multi}. %,rezende2016unsupervised}
These models are reliant on additional data, such as depth information or mesh models, to perform their target tasks, however,
and cannot be trained using images alone.
%%
%% CLAMPING
Other works have introduced a clamping strategy to enforce a specific organizational structure in the latent space~\cite{reed2014learning,kulkarni2015deep};
these networks require extremely detailed labels for supervision, such as the graphics code parameters used to create each example, and are therefore very difficult to implement for more general tasks (e.g. training with real images).
\cite{zhou2016view} have proposed the appearance flow network (AFN) designed specifically for the prediction of rotated viewpoints of objects from images.
This framework also relies on geometric concepts unique to rotation and is not generalizable to other inference tasks.
The conditional variational autoencoder (CVAE) incorporates conditioning information into the standard variational autoencoder (VAE) framework~\cite{kingma2013auto}
and is capable of synthesizing specified attribute changes in an identity preserving manner~\cite{sohn2015learning,yan2016attribute2image}.
CVAE-GAN~\cite{bao2017cvae} further adds adversarial training to the CVAE framework
in order to improve the quality of generated predictions.
\cite{zhang2017age} have introduced the conditional adversarial autoencoder (CAAE) designed to model age progression/regression in human faces. %; this model is trained to map input images of faces into different target age classes in an identity-preserving manner.
This is achieved by concatenating conditioning information (\ie age) with the input's latent representation before proceeding to the decoding process.
The framework also includes an adaptive discriminator with conditional information passed using a resize/concatenate procedure.
\cite{isola2017image} have proposed Pix2Pix as a general-purpose image-to-image translation network capable of synthesizing views from a single image.
The IterGAN model introduced by~\cite{galama2018iterative} is also designed to synthesize novel views from a single image, with a specific emphasis on the synthesis of rotated views of objects in small, iterative steps.
To the best of our knowledge, all existing conditional generative models designed for inference use fixed hidden layers and concatenate conditioning information directly with latent representations; in contrast to these existing methods, the proposed model incorporates conditioning information by defining dedicated, transformation-specific convolutional layers at the latent level.
    This conditioning framework allows the network to synthesize
    multiple transformed views from a single input,
    while retaining a fully-convolutional structure
    which avoids the
    dense connections used in existing inference-based conditional models.
    Most significantly, the proposed LTNN framework is shown to
    outperform state-of-the-art models in a diverse range of view synthesis tasks,
    while requiring substantially less FLOPs for inference than other conditional generative models (see Tables~\ref{table:aloi_results}~\&~\ref{table:chair_results}).

%% LATENT TRANSFORMATION NEURAL NETWORK
\section{Latent Transformation Neural Network}
In this section, we introduce the methods used to define the proposed LTNN model.
We first give a brief overview of the LTNN network structure.
We then detail how conditional transformation unit mappings are defined and trained to operate on the latent space,
followed by a description of the conditional discriminator unit implementation
and the network loss function used to guide the training process.
Lastly, we describe the task-division framework used for the decoding process.

The basic workflow of the proposed model is as follows:%outlined by the following steps:
\begin{enumerate}
\item Encode the input image $x$ to a latent representation $\, l_x \, = \, \operatorname{Encode}(x)$.
\item Use conditioning information $k$ to select conditional, convolutional filter weights $\omega_k$. % corresponding to the specified transformation.
\item Map the latent representation $\, l_x \, $ to $ \, \, \widehat{l}_{y_k}  =  \Phi_k(l_x)  =  \operatorname{conv}(l_x, \omega_k)$, an approximation of the encoded latent representation $l_{y_k}$ of the specified target image $y_k$.
\item Decode $\widehat{l}_{y_k}$ to obtain a coarse pixel value map and a refinement map.
\item Scale the channels of the pixel value map by the RGB balance parameters and take the Hadamard product with the refinement map to obtain the final prediction $\widehat{y}_k$.
  %%
  %\item The discriminator is passed real images $y_k$ as well as generated images $\widehat{y}_k$ and is trained to differentiate between the two image classes.
\item Pass real images $y_k$ as well as generated images $\widehat{y}_k$ to the discriminator, and use the conditioning information to select the discriminator's conditional filter weights $\overline{\omega}_k$.
\item Compute loss and update weights using ADAM optimization and backpropagation.
\end{enumerate}

\begin{algorithm}[t]
  \renewcommand{\thealgorithm}{}
  \caption{{\bf{LTNN Training Procedure}}}
  \label{training_procedure}

  \vspace{0.075in}
  \hspace{-0.045in} \textbf{Provide:}  Labeled dataset $\big\{\big(x,\{y_k\}_{k\in\mathcal{T}}\big)\big\}$ with target transformations indexed by a fixed set $\mathcal{T}$, encoder weights $\theta_E$, decoder weights $\theta_D$, RGB balance parameters $\{\theta_{R},\theta_{G},\theta_{B}\}$, conditional transformation unit weights $\{\omega_{k}\}_{k\in\mathcal{T}}$, discriminator $\mathcal{D}$ with standard weights $\theta_\mathcal{D}$ and conditionally selected weights $\{\overline{\omega}_{k}\}_{k\in\mathcal{T}}$, and loss function hyperparameters $\gamma, \rho, \lambda, \kappa$ corresponding to the smoothness, reconstruction, adversarial, and consistency loss terms, respectively.  The specific loss function components are defined in detail in Equations \ref{eq:adv2} - \ref{eq:consist} in Section \ref{sec:loss}.

  \begin{algorithmic}[1] % The number tells where the line numbering should start

    \vspace{0.075in}
    \Procedure{Train}{ } %\Comment{The g.c.d. of a and b}
    \State $x \, , \{y_k\}_{k\in\mathcal{T}} \,  = \, \operatorname{get\_train\_batch}()$ \Comment{Sample input and targets from training set}
    \State $l_x  \, = \,  \operatorname{Encode}[ \, x \, ]$     \Comment{Encoding of original input image}
    \For{$k$ in $\mathcal{T}$}
    \State $l_{y_k}  \, = \,  \operatorname{Encode}[ \, y_k \, ]$    \Comment{True encoding of specified target image}
    \State $\widehat{l}_{y_k} \, = \, \operatorname{conv}(l_x,\omega_k)$  \Comment{Approximate encoding of target with CTU}
    \State $\widehat{y}^{\,value}_{k} \, , \, \widehat{y}^{\,refine}_{k} \, = \, \operatorname{Decode}[\,\,\widehat{l}_{y_k}  \,] $  \Comment{Compute RGB value and refinement maps}
    \State $\widehat{y}_k \, = \, \big[ \, \theta_C \,\cdot\, \widehat{y}^{\,value}_{k,C} \odot\, \widehat{y}^{\,refine}_{k,C} \, \big]_{C\in\{\!R,G,B\!\}} $  \Comment{Assemble final network prediction for target}
    \State
    \State \# Update encoder, decoder, RGB, and CTU weights
    \State $\mathcal{L}_{adv} \, = \, -\log(\mathcal{D}(\widehat{y}_k, \overline{\omega}_k))$
    \State $\mathcal{L}_{guide} \, = \,  \gamma \cdot  \mathcal{L}_{smooth}(\widehat{y}_k) + \rho \cdot \mathcal{L}_{recon}(\widehat{y}_k,y_k) $
    \State $\mathcal{L}_{consist} \, = \,   \|\widehat{l}_{y_k} - l_{y_k}\|_{1}$
    \State $\mathcal{L} \, = \, \lambda \cdot \mathcal{L}_{adv} +  \mathcal{L}_{guide} + \kappa \cdot \mathcal{L}_{consist}$
    \For{$\theta$ in $\{\theta_E, \theta_D, \theta_{R}, \theta_{G}, \theta_{B}, \omega_{k}\}$}
    \State $\theta \, = \,  \theta - \nabla_{\theta} \mathcal{L}$
    \EndFor
    \State
    \State \# Update discriminator and CDU weights
    \State $\mathcal{L}^\mathcal{D}_{adv} \, = \, -\log(\mathcal{D}(y_k, \overline{\omega}_k)) -\log(1 \, - \, \mathcal{D}(\widehat{y}_k, \overline{\omega}_k))$
    \For{$\theta$ in $\{\theta_\mathcal{D}, \overline{\omega}_{k}\}$}
    \State $\theta \, = \,  \theta - \nabla_{\theta} \mathcal{L}^\mathcal{D}_{adv}$
    \EndFor
    \EndFor
    \EndProcedure
  \end{algorithmic}
\end{algorithm}

%%%%%%%%% CONDITIONAL TRANSFORMATION UNIT %%%%%%%%%

%\section{Preliminaries}
\subsection{Conditional transformation unit} \label{sec:ctu}
\begin{figure}[h]
  \begin{center}
    \includegraphics[width=1.\linewidth,keepaspectratio]{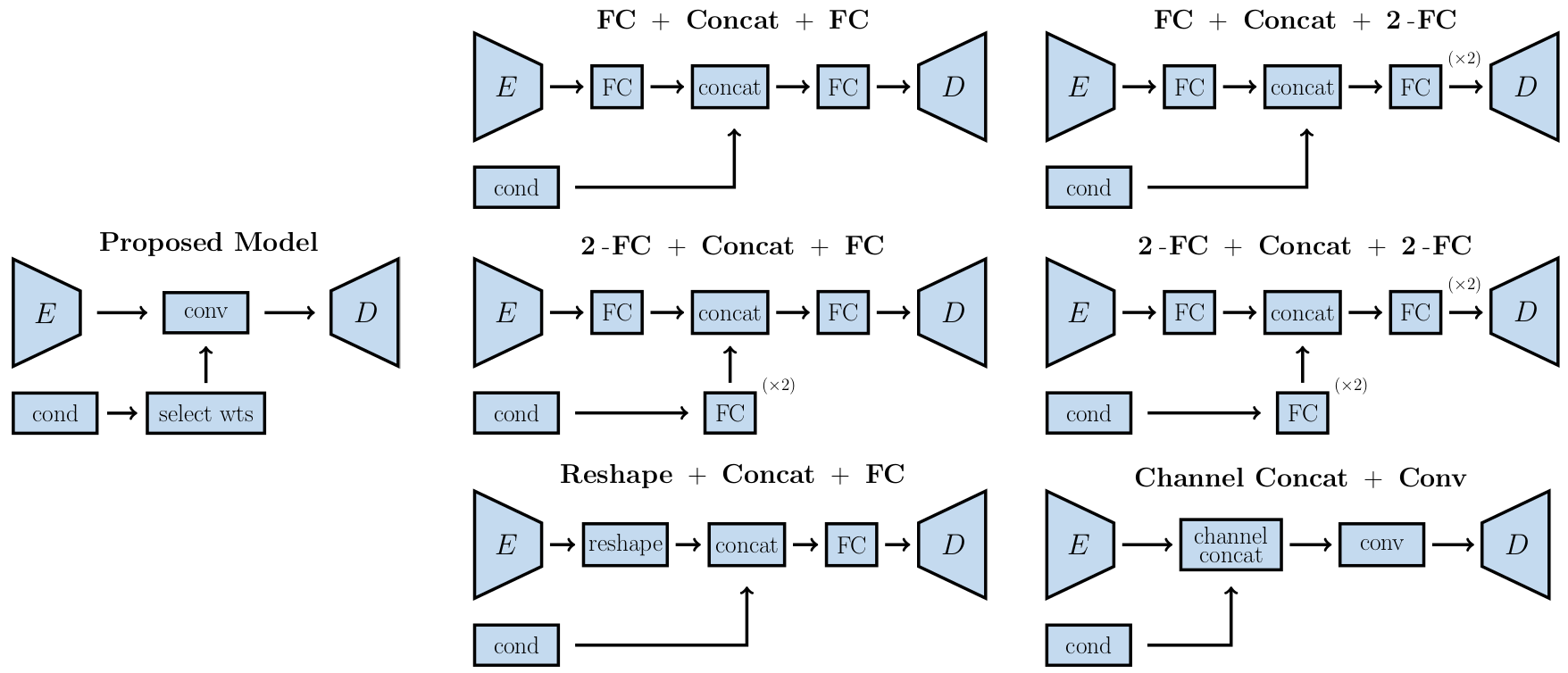}
  \end{center}
  \caption{Selected methods for incorporating conditioning information;  the proposed LTNN method is illustrated on the left, and six conventional alternatives are shown to the right.} %; the `concat' layers all implicitly include fully-connected layers before and after concatenation in order to connect with the convolutional layers of the  encoder and decoder. }
  \label{fig:cond_diagrams}
\end{figure}

Generative models have frequently been designed to explicitly disentangle the latent space in order to enable high-level attribute modification through linear, latent space interpolation.  %A guiding motivation of this work comes from the intuition that this linear orginizational structure
This linear latent structure is imposed by design decisions, however, and may not be the most natural way for a network to internalize features of the data distribution.
%To lift this modeling constraint, we propose an additional non-linear network layer which acts on the latent space itself; we refer to these latent space layers as {\emph{conditional transformation units}}.
%To lift this modeling constraint, s
Several approaches have been proposed which include nonlinear layers for processing conditioning information at the latent space level.  %Conventionally, these networks use dense connections to the latent space and incorporate the processed conditioning information via concatenation; six of these conventional conditional network designs are illustrated in Figure \ref{fig:cond_diagrams} along with the proposed LTNN network design for incorporating conditioning information.
%%% ADDED
In these conventional conditional generative frameworks, conditioning information is introduced by combining features extracted from the input with features extracted from the conditioning information (often using dense connection layers);
these features are typically combined using standard vector concatenation, although some have opted to use channel concatenation~\cite{zhang2017age,bao2017cvae}.
Six of these conventional conditional network designs are illustrated in Figure \ref{fig:cond_diagrams} along with the proposed LTNN network design for incorporating conditioning information.
%in most cases these features {\rd{flattened}} and are combined by vector concatenation, while {\rd{REFERENCE}} propose the use of channel concatenation.

Rather than directly concatenating conditioning information,
we propose using a conditional transformation unit (CTU),
consisting of a collection of distinct convolutional mappings in the network's latent space;
conditioning information is then used to select which collection of weights, i.e. which CTU mapping, should be used
in the convolutional layer to perform a specified transformation.
For view point estimation, there is an independent CTU per viewpoint.
Each CTU mapping maintains its own collection of convolutional filter weights and uses Swish activations~\cite{ramachandran2017swish}.
The filter weights and Swish parameters of each CTU mapping are selectively updated by controlling the gradient flow based on the conditioning information provided. %and target output associated with the mapping.
The CTU mappings are trained to transform the encoded, latent space representation of the network's input in a manner which produces high-level view or attribute changes upon decoding.
This is accomplished by introducing a {\emph{consistency}} term into the loss function which is minimized precisely when the CTU mappings behave as depicted in Figure~\ref{fig:manifold}.
In this way, different angles of view, light directions, and deformations, for example, can be %generated
synthesized from a single input image.

\subsection{Discriminator and loss function}
\label{sec:loss}

The discriminator used in the adversarial training process is also passed conditioning information which specifies the transformation which the model has attempted to make.
The conditional discriminator unit (CDU), consisting of convolutional layers with modular weights similar to the CTU, is trained to specifically identify unrealistic artifacts which are being produced by the corresponding conditional transformation unit mappings.
For view point estimation, there is an independent CDU per viewpoint.

The proposed model uses the adversarial loss as the primary loss component.
The discriminator, $\mathcal{D}$, is trained using the adversarial loss term $\mathcal{L}^\mathcal{D}_{adv}$ defined below in Equation \ref{eq:adv2}.
Additional loss terms  corresponding to structural reconstruction, smoothness \cite{jason2016back}, and a notion of consistency,
are also used for training the encoder/decoder:
\begin{flalign}
  & \mathcal{L}^\mathcal{D}_{adv} \, \hspace{0.19in}  =  \, \, - \log \mathcal{D}(y_k, \overline{\omega}_k) \, - \, \log \big(1 - \mathcal{D}(\widehat{y}_k, \overline{\omega}_k) \big) \label{eq:adv2}\\
  & \mathcal{L}_{adv} \, \hspace{0.19in}  =  \, \, - \log \mathcal{D}(\widehat{y}_k, \overline{\omega}_k) \label{eq:adv1}\\
  & \mathcal{L}_{recon} \, \hspace{0.085in} = \, \, \| \, \widehat{y}_k \, - \, y_k \, \|_2^2 \label{eq:recon}\\
  & \mathcal{L}_{smooth} \, = \, \, 1/8 \,\cdot \!\!\!\!  \sum_{i\in\{0,\pm1\}} \, \sum_{j\in\{0,\pm1\}} \,  \big\| \, \widehat{y}_k \, - \, \tau_{i,j}\widehat{y}_k \, \big\|_1 \label{eq:smooth} \\
  & \mathcal{L}_{consist} \, \hspace{0.01in} = \, \, \big\| \, \Phi_k(\operatorname{Encode}[x]) \, - \, \operatorname{Encode}[y_k] \, \big\|_1 \label{eq:consist}
\end{flalign}
\noindent where $y_k$ is the modified target image corresponding to an input $x$,
\, $\overline{\omega}_k$ are the weights of the CDU mapping corresponding to the $k^{th}$ transformation,
$\Phi_k$ is the CTU mapping for the $k^{th}$ transformation,
$\widehat{y}_k \,  = \, \operatorname{Decode}\big( \Phi_k\big(\operatorname{Encode}[x]\big) \big)$ is the network prediction,
and $\tau_{i,j}$ is the two-dimensional, discrete shift operator.
The final loss function for the encoder and decoder components is given by: % the weighted sum:
\begin{equation} \label{eq:total_loss}
  \mathcal{L} \, \, = \, \, \lambda\cdot \mathcal{L}_{adv}  \, + \, \rho\cdot\mathcal{L}_{recon} \, + \, \gamma \cdot \mathcal{L}_{smooth} \,+ \, \kappa \cdot \mathcal{L}_{consist}
\end{equation}
\noindent with hyperparameters typically selected so that  $\lambda, \rho \gg  \gamma, \kappa$.  %The smooth loss is %added as a regularization term
%included to encourage smooth predictions.
The consistency loss is designed to guide the CTU mappings toward approximations of the latent space mappings which connect the latent representations of input images and target images as depicted in Figure \ref{fig:manifold}.
In particular, the consistency term enforces the condition that the transformed encoding, $\widehat{l}_{y_k} = \Phi_k(\mbox{Encode}[x])$, approximates the encoding of the $k^{th}$ target image, $l_{y_k} = \mbox{Encode}[y_k]$, during the training process.
%The discriminator weights are trained using exclusively the adversarial loss term $\mathcal{L}^\mathcal{D}_{adv}$ defined above in Equation \ref{eq:adv2}.

\subsection{Task separation / Task-divided decoder}
\label{sec:task-divide}
%We divide the decoding process into three tasks: estimating the certainty %probability
The decoding process has been divided into three tasks: estimating the refinement map, pixel-values, % estimation,
and RGB color balance of the dataset.
We have found
%Experiments have shown
this decoupled framework for estimation helps the network  % local/global
converge to better minima to produce sharp, realistic outputs.
%In the decoding process,  convolutional layers with bilinear interpolation are used to upsample the low resolution latent information.  %The last component of the upsampling process consists of two distinct transpose convolutional layers used for task separation; one layer is allocated for predicting the refinement map while the other is trained to predict pixel-values.
%The decoder is designed to produce a value map, which serves as an approximation to the desired output, along with a refinement map. %, which is aimed to model the confidence of the network has with regard to the value map it has produced.
The decoding process begins with a series of convolutional layers followed by bilinear interpolation to upsample the low resolution latent information.
The last component of the decoder's upsampling process consists of two distinct transpose convolutional layers used for task separation; one layer is allocated for predicting the refinement map while the other is trained to predict pixel-values.
The refinement map layer incorporates a Sigmoid activation function which outputs scaling factors intended to refine the coarse pixel value estimations.
RGB balance parameters, consisting of three trainable variables, are used as weights for balancing the color channels of the pixel value map.
The Hadamard product of the refinement map and the RGB-rescaled value map serves as the network's final output: %approximation to the desired output.
\begin{equation} \label{eq:divide}
  \widehat{y}  \, =  \, \left[ \, \widehat{y}_{R}, \, \widehat{y}_{G}, \, \widehat{y}_{B} \, \right]  \hspace{0.075in}\mbox{where}\hspace{0.1in}  \widehat{y}_{C} \, = \, \theta_C \, \cdot \hspace{0.03in} \widehat{y}^{\, \, value}_C \, \odot \, \widehat{y}^{\, \, refine}_C  \hspace{0.075in}\mbox{for}\hspace{0.075in} C \hspace{0.01in} \in \hspace{0.01in} \{R, G, B\}
\end{equation}

In this way, the network has the capacity to mask values which lie outside of the target object (\ie by setting refinement map values to zero) which allows the value map to focus on the object itself during the training process.
Experimental results show that the refinement maps learn to produce masks which closely resemble the target objects' shapes and have sharp drop-offs along the %objects'
boundaries. % (see Figure~\ref{fig:refine_map}).
%%
%{\rd{In addition to masking extraneous pixels, these refinement maps have been shown to apply local color balancing by, for example, filtering out the green and blue channels near lips when applied to human faces (see Figure~\ref{fig:refine_map}). [IF LOW ON SPACE, MAY OMIT]}}

%% OMITTED
%{\rd{It has been observed that the training process proceeds sequentially through two distinct phases;  the model first learns to accurately approximate the target shape and subsequently begins to focus on pixel value-estimation.}}

%% HAND DIAGRAM FROM MAIN TEXT
\begin{figure}[h]
  \begin{center}
    \includegraphics[width=1.0\linewidth]{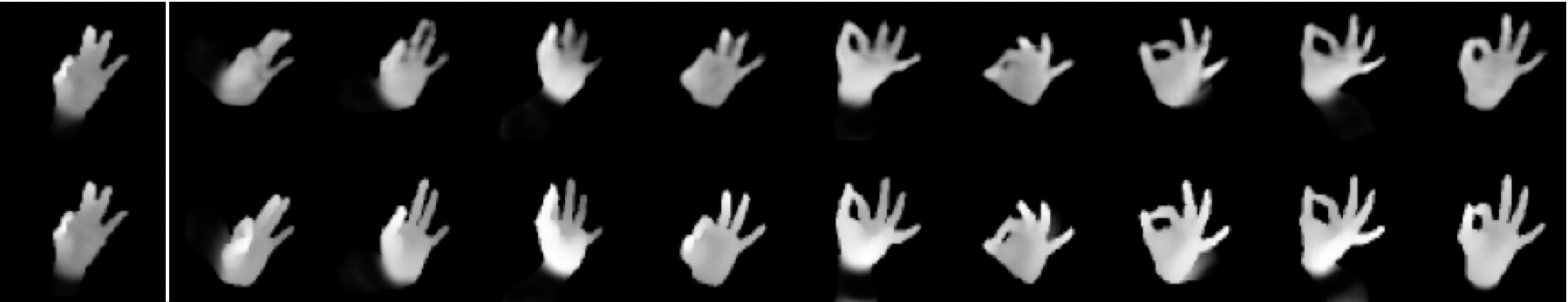}
  \end{center}
  \caption{ Comparison of CVAE-GAN (top) with proposed LTNN model (bottom) using the noisy NYU hand dataset~\cite{tompson2014real}.  The input depth-map hand pose image is shown to the far left, followed by the network predictions for 9 synthesized view points. The views synthesized using LTNN are seen to be sharper and also yield higher accuracy for pose estimation (see Figure~\ref{fig:face_real}).} %The left-most column shows the given input image view point of the hand pose; the following right 9 images are the estimated views. The top row images are generated from CVAE-GAN and the bottom images are estimated from our method. The images from our method show better results.}
  \label{fig:hand_qual}
\end{figure}

\begin{figure}[h]
  \centering
    \includegraphics[width=1.0\linewidth]{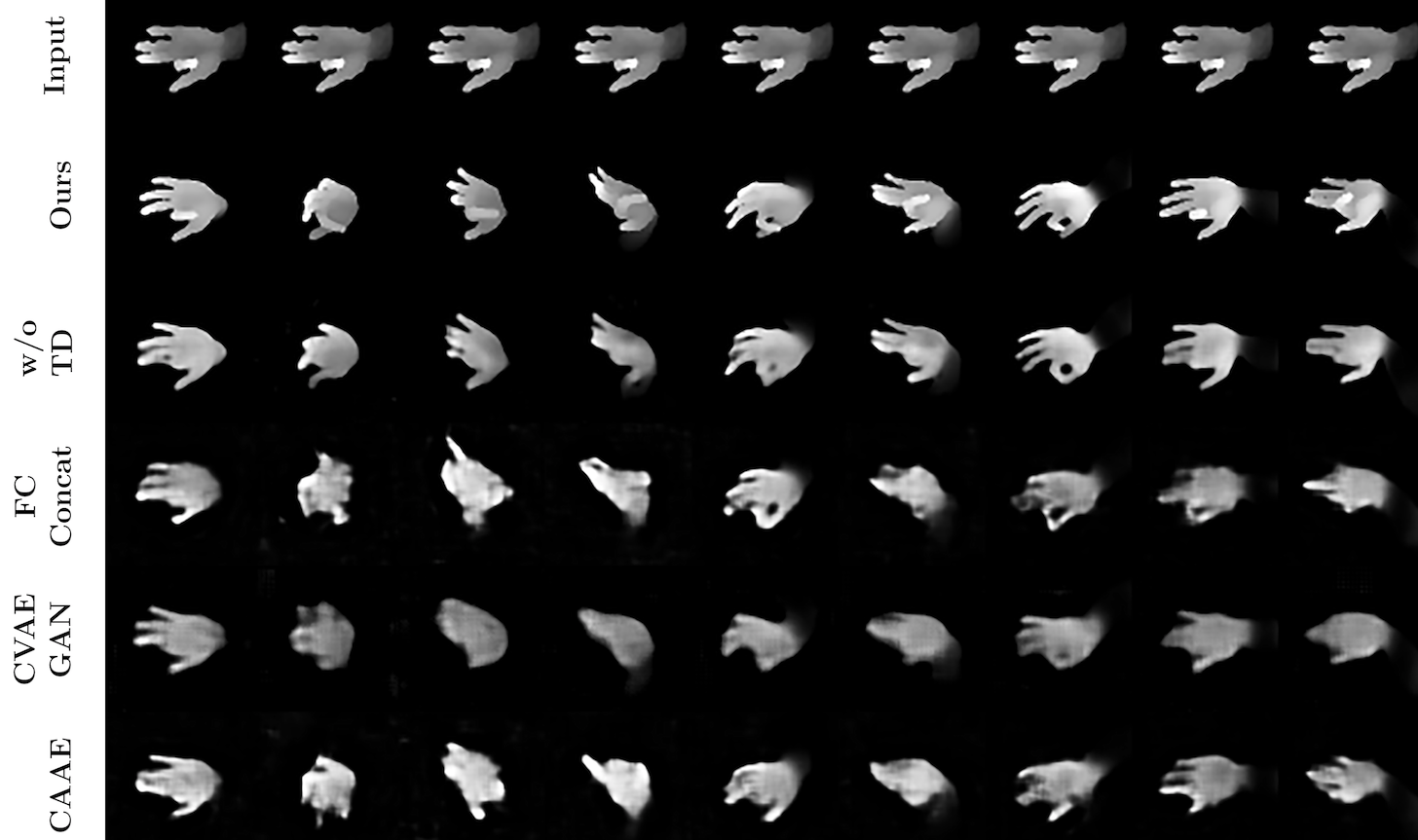}
  \caption{Qualitative evaluation for multi-view reconstruction of hand depth maps using the NYU dataset.}
  \label{fig:hand_qual_main}
\end{figure}

%%%%%%%%% EXPERIMENTS %%%%%%%%%
\section{Experiments and results}
%We conduct evaluations on a diverse range of datasets including both rigid and non-rigid objects.
%%
%%
To show the generality of our method, we have conducted a series of diverse experiments:
(i)
hand pose estimation using a synthetic training set and real NYU hand depth image data~\cite{tompson2014real} for testing,
(ii)
synthesis of rotated views of rigid objects using the real ALOI dataset~\cite{geusebroek2005amsterdam} and synthetic 3D chair dataset~\cite{shapenet2015},
(iii)
synthesis of rotated views using a real face dataset~\cite{fransens2005parametric},
and
(iv)
the modification of a diverse range of attributes on a synthetic face dataset~\cite{bfm09}.
For each experiment, we have trained the
models using 80\% of the datasets.
%%
%%We search hyper parameters for the network with syentatic datasets only.
%%
Since ground truth target depth images were not available for %the experiment on
the real hand dataset,
an indirect metric has been used to quantitatively evaluate the model as described in Section~\ref{sec:nonrigid}.
Ground truth data was available for all other experiments, and
models were evaluated directly using the
%standard
$L_1$ mean pixel-wise error and the structural similarity index measure (SSIM)~\cite{park2017transformation,mathieu2015deep} (the masked pixel-wise error $L_1^M$~\cite{galama2018iterative} was used in place of the $L_1$ error for the ALOI experiment).
%% , masked pixel-wise error $L_1^M$~\cite{galama2018iterative}

\begin{figure}[ht]
  \begin{center}
    \begin{tabular}{c c}
      \includegraphics[width=0.44\linewidth]{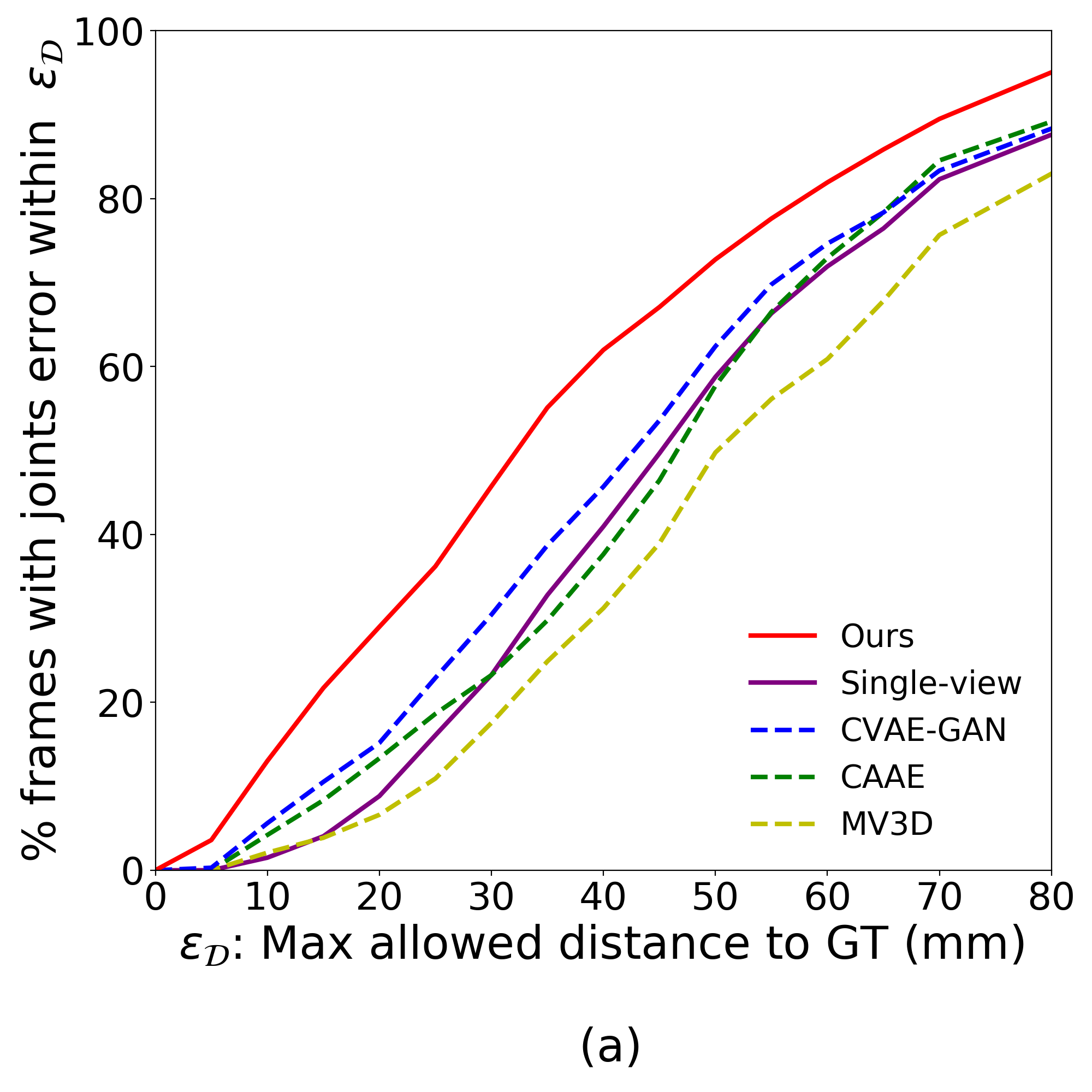}
      &
      \includegraphics[width=0.44\linewidth]{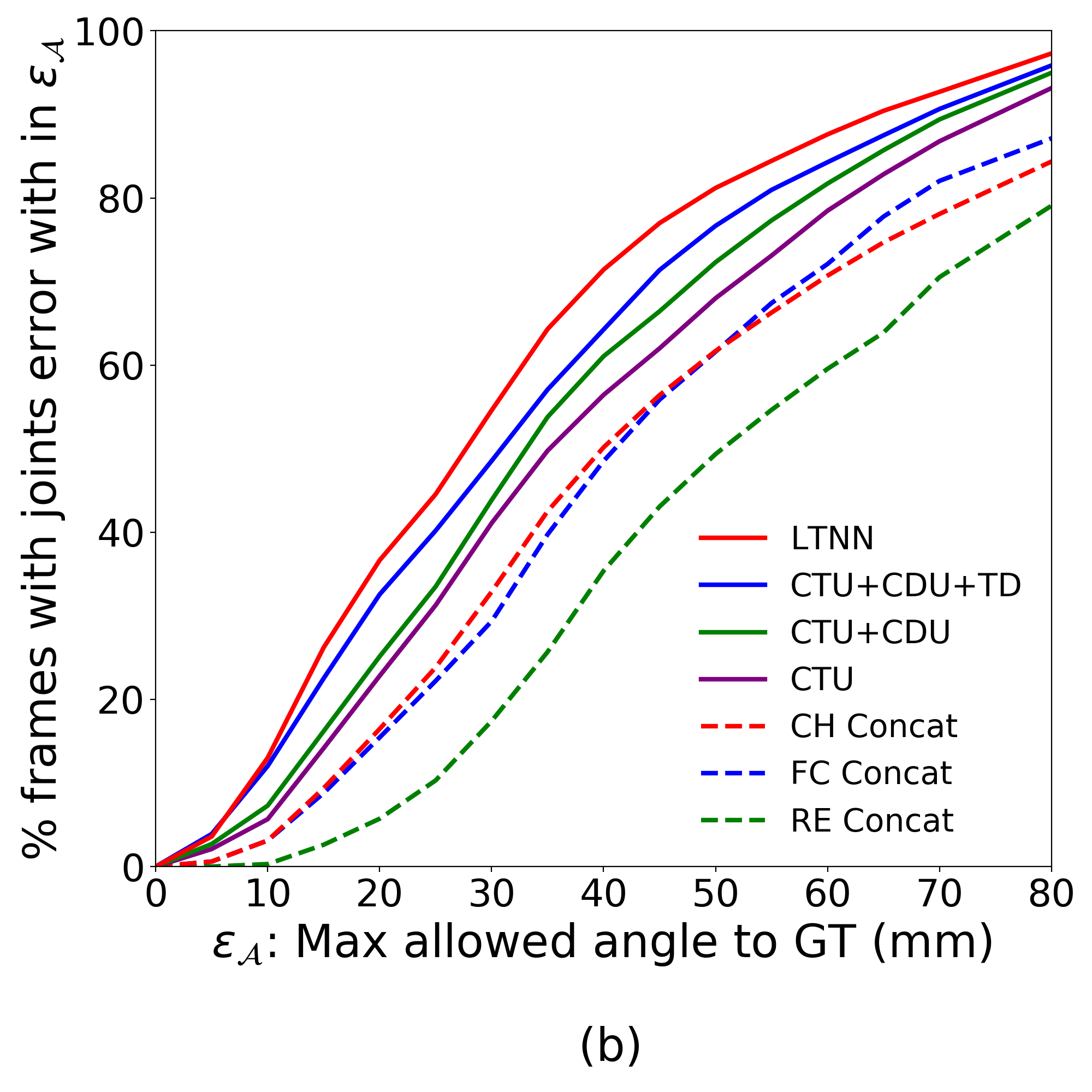}
    \end{tabular}
  \end{center}
  \caption{Quantitative evaluation for multi-view hand synthesis.
    (a) Evaluation with state-of-the-art methods using the real NYU dataset.
    (b) LTNN ablation results and comparison with alternative conditioning frameworks
    %with self-generated baselines
    using synthetic hand dataset. % for validation before testing on NYU.
    Our models: conditional transformation unit (CTU), conditional discriminator unit (CDU), task-divide and RGB balance parameters (TD), and LTNN consisting of all previous components along with consistency loss. Alternative concatenation methods: channel-wise concatenation (CH Concat), fully connected concatenation (FC Concat), and reshape concatenation (RE Concat).}
  \label{fig:hand_cdf}
\end{figure}

To evaluate the proposed framework with existing works, two comparison groups have been formed:
conditional inference models (CVAE-GAN, CVAE, and CAAE) with comparable encoder/decoder structures for comparison on experiments with non-rigid objects, and
view synthesis models (MV3D~\cite{tatarchenko2016multi}, IterGAN, Pix2Pix, AFN~\cite{zhou2016view}, and TVSN~\cite{park2017transformation}) for comparison on experiments with rigid objects.
Additional experiments have been performed to compare the proposed CTU conditioning method with other conventional concatenation methods (see Figure~\ref{fig:cond_diagrams}); results are shown in Figure~\ref{fig:hand_cdf}.
%%

%For ablation study on each proposed methods, we stacked each method on top of the base model CTU. We added in following order: task-divide and RGB balance parameters (TD), conditional discriminate units (CDU), consistance loss (CON).

%% MULTI-VIEW HAND GRAPHS

%% REAL FACE GRAPH
\begin{figure}[ht]
   \begin{center}
    \begin{tabular}{cc}
      \includegraphics[width=0.47\linewidth]{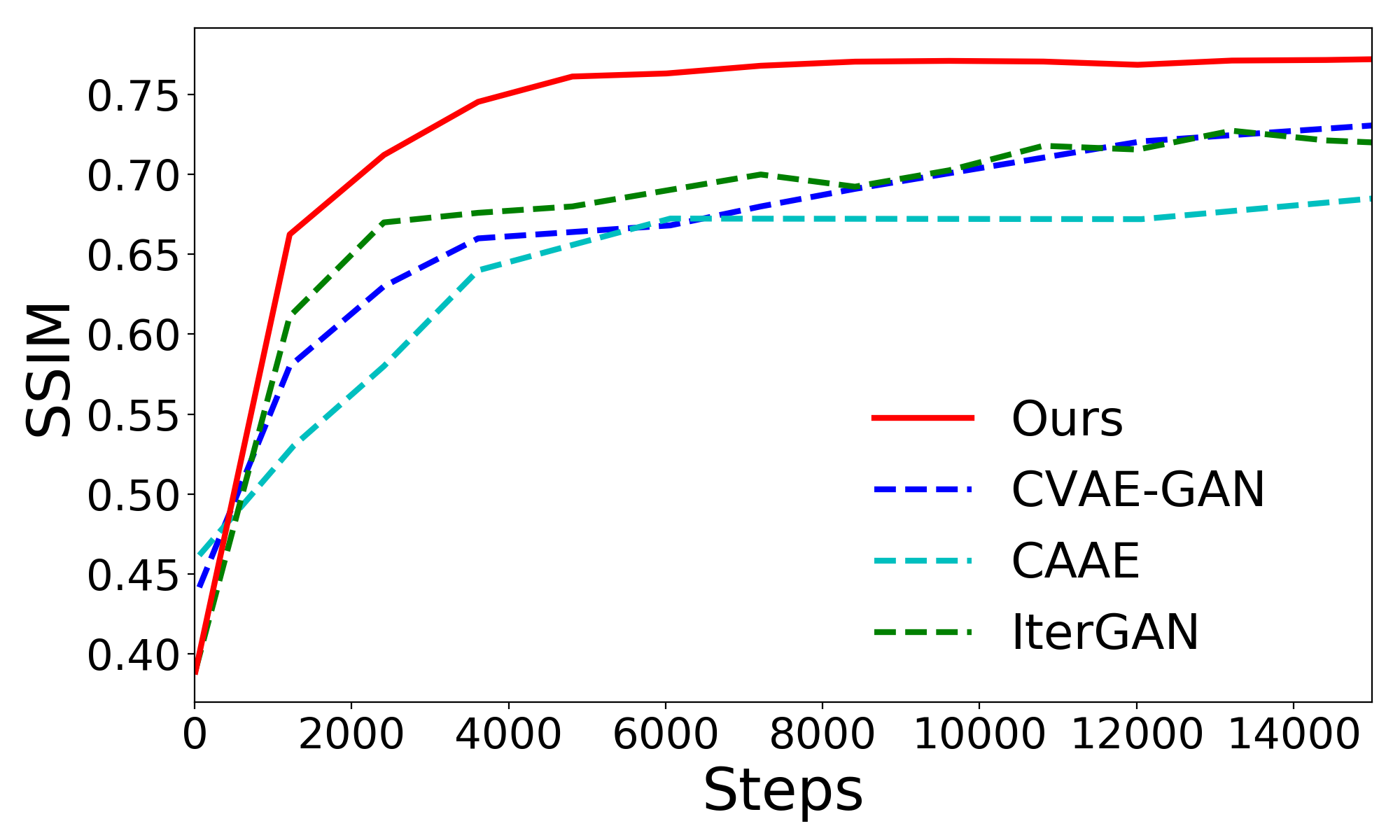}
      &
       \includegraphics[width=0.47\linewidth]{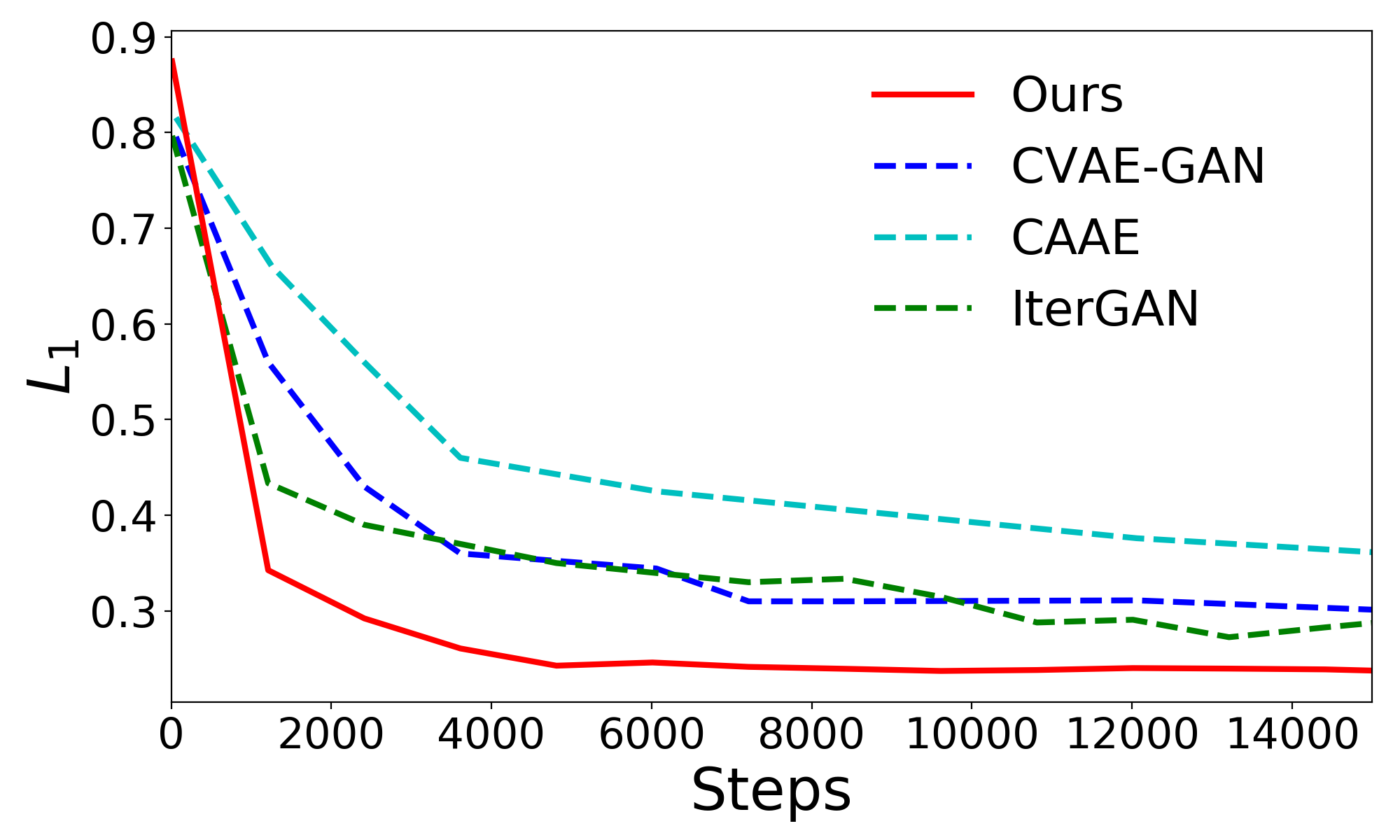}
    \end{tabular}
    \end{center}
   \caption{Quantitative comparison of model performances for experiment on the real face dataset.}%; SSIM (left) and $L_{1}$ (right).}%; CVAE-GAN, CAAE, and IterGAN.}
  \label{fig:face_real}
\end{figure}

\begin{figure}[h]
    \begin{center}
    \includegraphics[width=1.0\linewidth]{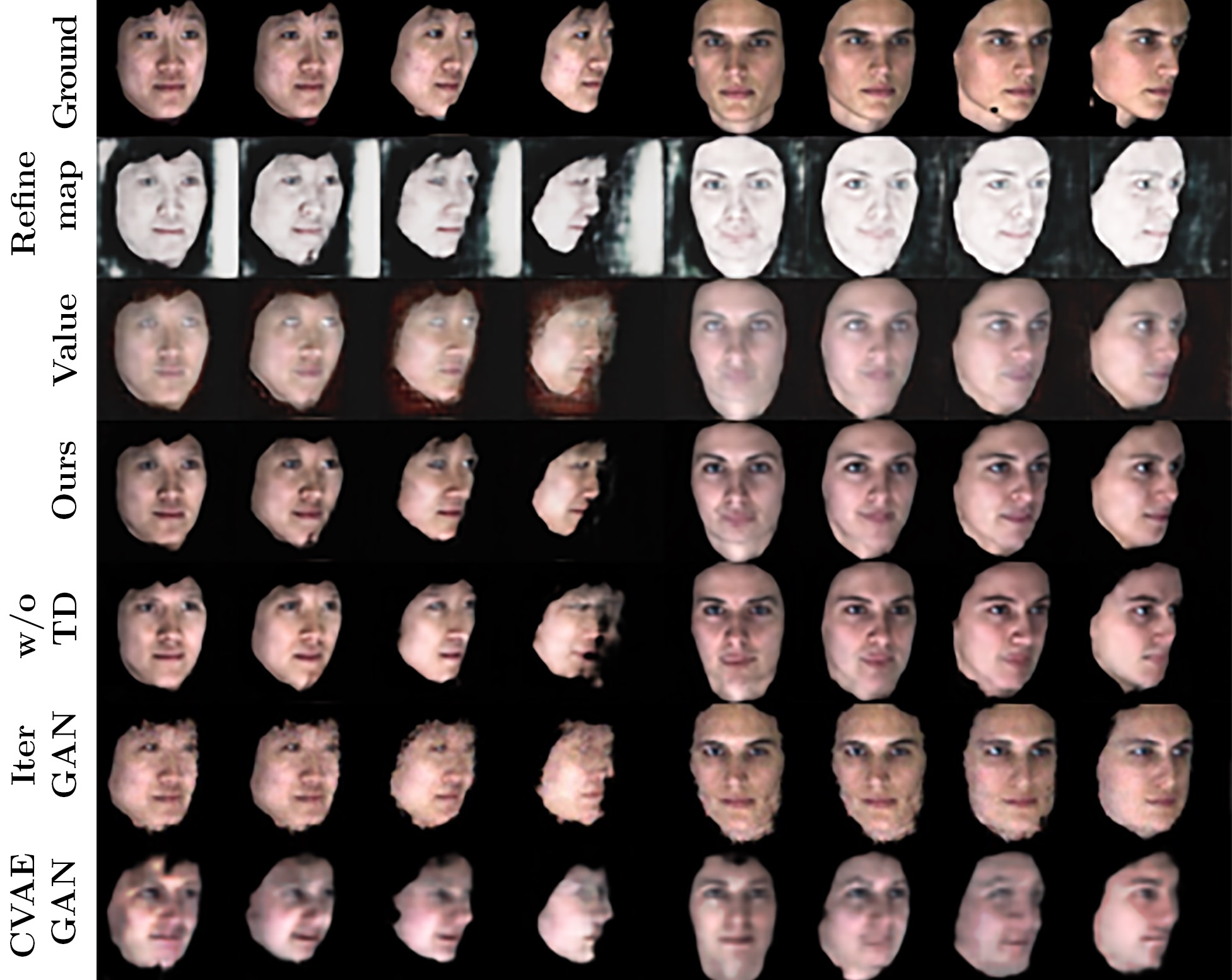}
    \end{center}
  \caption{Qualitative evaluation for multi-view reconstruction of real face using the stereo face dataset~\cite{fransens2005parametric}.}
  \label{fig:face_qual}
\end{figure}

%%% IMPLEMENTATION SECTION WAS HERE
\subsection{Experiment on non-rigid objects}
\label{sec:nonrigid}
%{\textbf{Hand pose view syentisizing:}
\textbf{Hand pose experiment:}
Since ground truth predictions for the real NYU hand dataset were not available,
the LTNN model has been trained using a synthetic dataset generated using 3D mesh hand models.
The NYU dataset does, however, provide ground truth coordinates for the input hand pose;
using this we were able to indirectly evaluate the performance of the model
by assessing the accuracy of a hand pose estimation method
using the  network's multi-view predictions as input.
More specifically, the LTNN model was trained to generate 9 different views which were then fed into the pose estimation network from~\cite{choi2017learning} (also trained using the synthetic dataset).

A comparison of the quantitative hand pose estimation results is provided in
Figure~\ref{fig:hand_cdf} where the proposed LTNN framework is seen to provide a substantial
improvement over existing methods;
qualitative results are also available in Figure~\ref{fig:hand_qual}.
%%
%Ground truth depth images for the generated views were not available, however
%Compare with other existing methods and each of our propoesed methods significantly help increasing the accuracy.
%%
With regard to real-time applications, the proposed model runs at 114 fps without batching and at 1975 fps when applied to a mini-batch of size 128  (using a single TITAN Xp GPU and an Intel i7-6850K CPU).

\textbf{Real face experiment:}
%For real face experiment we used
The stereo face database~\cite{fransens2005parametric}, consisting of images of 100 individuals from 10 different viewpoints, was used for experiments with real faces;
these faces were segmented using the method of ~\cite{nirkin2018face} and then cropped and centered to form the final dataset.
The LTNN model was trained to synthesize images of input faces corresponding to three consecutive horizontal rotations.
As shown in Figure~\ref{fig:face_real}, our method significantly outperforms the CVAE-GAN, CAAE, and IterGAN models in both the $L_1$ and SSIM metrics. % and abel to generate realistic faces. There are qualitive results at the appendix.
%%
%We first segment faces with the face segmentation method~\cite{nirkin2018face} and manually filterout failure cases. We paired 4 continuous view points for training and testing. As shown in the two plots at the Figure~\ref{fig:face_real}, our method outperforms CVAE-GAN, CAAE, and IterGAN in both L1 and SSIM metrics and abel to generate realistic faces. There are qualitive results at the appendix.

%TODO Do triplet joint position regression for the table results or CDF

\subsection{Experiment on rigid objects}
%TODO Require 128 resolution training and if have a time deformation

%To illustrate the proposed model's ability to interpret th three-dimensional structure of rigid objects, we have conducted an experiment  \ang{360} viewpoint estimation on the chair dataset.
%\textbf{Real 1000 objects:}

\begin{figure}[h]
    \begin{center}
      \begin{tabular}{c}
    \includegraphics[width=1.0\linewidth]{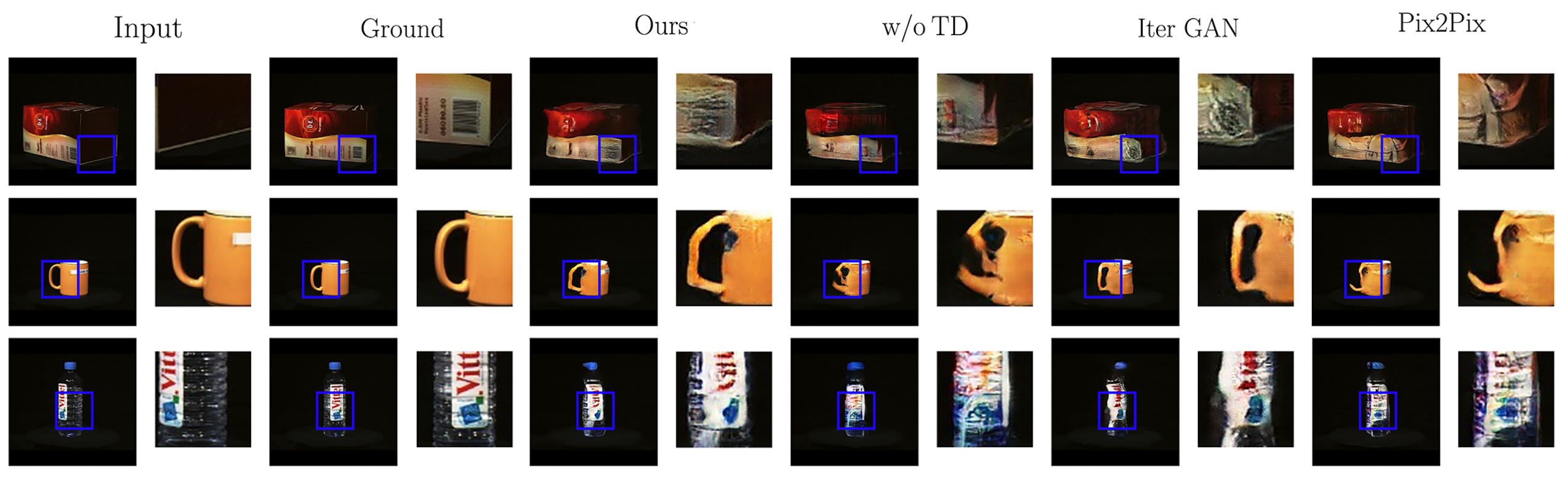}
      \end{tabular}
    \end{center}
    \caption{Unseen objects qualitative evaluation for ALOI dataset.}
  \label{fig:unseen}
\end{figure}

\textbf{Real object experiment:}
The ALOI dataset~\cite{geusebroek2005amsterdam}, consisting of images of 1000 real objects viewed from 72 rotated angles (covering one full \ang{360} rotation), has been used for experiments on real objects.
%%
%{\rd{Following the experimental setup of IterGAN...}}
%%
As shown in Table~\ref{table:aloi_results} and in Figure~\ref{fig:unseen}, our method outperforms other state-of-the-art methods with respect to the $L_{1}$ metric and achieves comparable SSIM metric scores.

%%
%The experiment proves that LTNN is robust on rigid-objects as well.
% We experiment our model on 1000 real objects with ALOI dataset~\cite{geusebroek2005amsterdam} which has 72 rotation view images covering \ang{360} angles per object.
% From the experiment results, our methods outperform in $L_{1}$ metric and compariabel resutls on SSIM metric with other state-of-the-art methods, as shown on the Table~\ref{table:objects_1}. The experiment proves that LTNN is robust on rigid-objects as well.
Of note is the fact that the LTNN framework is capable of effectively performing the specified rigid-object transformations
using only a single image as input,
whereas most state-of-the-art view synthesis methods require additional information which is not practical to obtain for real datasets.
For example, MV3D requires depth information and
TVSN requires 3D models to render visibility maps for training
%{\rd{and AFN requires view point transformation vector as input for testing and training [DOUBLE CHECK]}},
which is not available in the ALOI dataset.
%%
%However, our methods dosen't require 3D mesh model nor depth information for \ang{360} view syentisizing.

%% ALOI TABLE
\begin{table}[h]
  \begin{center}
\begin{tabular}{c c c c c}
  \hline
  Model& $L_{1}^{M}$~seen & $L_{1}^{M}$~unseen & SSIM~seen & SSIM~unseen\\
  \hline
  Ours   & \textbf{.138$\pm$.046} & \textbf{.221$\pm$.064} & \textbf{.927$\pm$.012} & .871$\pm$.031\\
  IterGAN& .147$\pm$.055 & .231$\pm$.094 & .918$\pm$.019 & \textbf{.875$\pm$.025}\\
  %AFN&.$\pm$.&.$\pm$.&.$\pm$.&.$\pm$.\\
  Pix2Pix& .210$\pm$.092 & .256$\pm$.100 & .914$\pm$.021 & .864$\pm$.041\\
  \hline
\end{tabular}
  \end{center}
%\end{minipage}
\caption{The result table for the experiment on the ALOI dataset.
}
\label{table:aloi_results}
\end{table}

\begin{figure}[h]
  \begin{center}
    \includegraphics[width=1.0\linewidth]{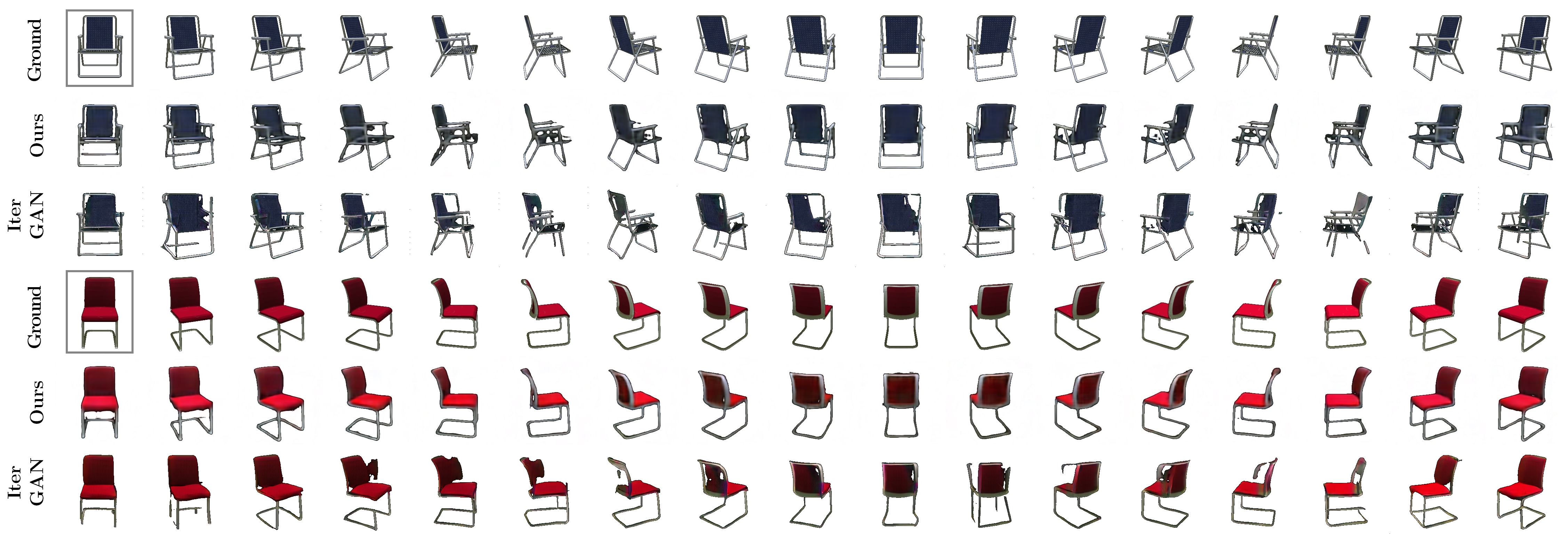}
  \end{center}
  \caption{  Generated \ang{360} views for chair dataset.  A single, gray-scale image of the chair at the far left (shown in box) is provided to the network as input.}
  \label{fig:chair_360}
\end{figure}

\textbf{3D chair experiment:}
We have tested our model's ability to perform \ang{360} view estimation on the chairs and compared the results with the other state-of-the-art methods.
The proposed model outperforms existing models specifically designed for the task of multi-view prediction
and require the least FLOPs for inference compared with all other methods (see Table~\ref{table:chair_results}).

\begin{table}[h]
\begin{center}
\begin{tabular}{c c c c c c}
  \hline
  Model& SSIM & $L_{1}$ & Parameters~for~train & Parameters~for~infer & GFLOPs~/~Image \\
  \hline
  Ours   & \textbf{.912} & \textbf{.217} & 65,438~K& \textbf{16,961~K} & \textbf{\; 2.183} \\
  IterGan & .865 & .351 & \textbf{59,951~K} & 57,182~K & 12.120 \\
  AFN    & .891 & .240 & 70,319~K & 70,319~K &  \; 2.671 \\
  TVSN   & .894 & .230 &60,224~K & 57,327~K&  \; 2.860 \\
  MV3D  & .895 & .248 & 69,657~K & 69,657~K & \; 3.056 \\
  \hline
\end{tabular}
\caption{Results for 3D chair \ang{360} view synthesis. The proposed method uses significantly less parameters during inference, requires the least FLOPs, and yields the fastest inference times. FLOP calculations correspond to inference for a single image with resolution 256$\times$256$\times$3.}
\label{table:chair_results}
%\end{minipage}
%\hspace{0.1cm}
%\begin{minipage}[b]{0.75\linewidth}\centering
\end{center}
%\centering
\end{table}

\begin{table}[h]
  \begin{center}
	\begin{tabular}{l c c c c c c c c}
	  \hline
      & \multicolumn{2}{c}{Elevation} & \multicolumn{2}{c}{Azimuth}  & \multicolumn{2}{c}{~~Light Direction~~} & \multicolumn{2}{c}{Age}  \\
	  \hline
	  Model& ~~SSIM~~ & ~~$L_{1}$~~ & ~~SSIM~~ & ~~$L_{1}$~~ &  ~~SSIM~~ & ~~$L_{1}$~~ & ~~SSIM~~ & ~~$L_{1}$~~  \\
	  \hline
	  Ours
	  & \textbf{.923}  & \textbf{.107}
	  & \textbf{.923} & \textbf{.108 }
      & \textbf{.941}& .\textbf{093}
      & \textbf{.925}& \textbf{.102} \\
      CVAE-GAN
	  & .864& .158
	  & .863& .180
	  & .824& .209
	  & .848& .166\\
	  CVAE
	  & .799 & .166
	  & .812 & .157
	  & .806 & .209
	  & .795 & .173 \\
    \iffalse
	  CVAE\footnotemark
	  & .784 & .177
	  & .782 & .184
	  & .604 & .243
	  & .763 & .197 \\
    \fi
	  CAAE
	  & .777 & .175
	  & .521 & .338
	  & .856 & .270
	  & .751 & .207 \\
	  AAE
	  & .748 & .184
	  & .520 & .335
	  & .850 & .271
	  & .737 & .209 \\
	  \hline
	\end{tabular}
  \end{center}
  \caption{Results for simultaneous colorization and attribute modification on synthetic face dataset.}
  \label{table:face3d_selfcompare}
\end{table}

\subsection{Diverse attribute exploration with synthetic face data}
\label{sec:diverse}
%%
%A synthetic face dataset has been used to assess the LTTN model on a more diverse range of
%tasks than are available in real datasets.
%%
%To evaluate the proposed framework's performance with regard to attribute modification of faces, %four variations of
%To evaluate the proposed framework's performance,
To evaluate the proposed framework's performance on a more diverse range of attribute modification tasks, a synthetic face dataset and
five conditional generative models with comparable encoder/decoder structures to the LTNN model have been selected for comparison.
These models have been trained to synthesize discrete changes in elevation, azimuth, light direction, and age
from a single greyscale image;
results are shown in Table~\ref{table:face3d_selfcompare}.
Near continuous attribute modification is also possible within the proposed framework,
and distinct CTU mappings can be composed with one another to synthesize multiple modifications simultaneously.

%% CONCLUSION
\section{Conclusion}
In this work, we have introduced an effective, general framework for incorporating conditioning information into
inference-based generative models.
We have proposed a modular approach to incorporating conditioning information using CTUs and a consistency loss term, defined an efficient task-divided decoder setup for deconstructing the data generation process into manageable subtasks, and shown that a context-aware discriminator can be used to improve the performance of the adversarial training process.
The performance of this framework has been assessed on a diverse range of tasks and shown to outperform state-of-the-art methods.

\clearpage

\bibliographystyle{splncs}
\bibliography{egbib}
\end{document}